\documentclass[10pt,twocolumn,letterpaper]{article}

\usepackage[pagenumbers]{cvpr} 

\usepackage{graphicx}
\usepackage{amsmath}
\usepackage{amssymb}
\usepackage{booktabs}
\usepackage{bm}
\usepackage{array}
\usepackage{colortbl}  
\usepackage{multirow}
\usepackage{enumitem}
\usepackage{pifont}
\usepackage{comment}
\usepackage{mathtools} 
\usepackage{cuted}     
\usepackage{epigraph}
\usepackage{caption}

\definecolor{chrxcolor}{RGB}{1, 140, 116}

\definecolor{cvprblue}{rgb}{0.21,0.49,0.74}
\usepackage[pagebackref,breaklinks,colorlinks,allcolors=cvprblue]{hyperref}

\def\paperID{} 
\def\confName{CVPR}
\def\confYear{2026}

\begin{document}

\title{
ChangeBridge: Spatiotemporal Image Generation with Multimodal Controls \\ for Remote Sensing
\vspace{-6mm}
}


\author{
Zhenghui Zhao$^{1}$, Chen Wu$^{1\dagger}$, Xiangyong Cao$^{2\dagger}$,\\
Di Wang$^{3,4}$, Hongruixuan Chen$^{5}$, Datao Tang$^{2}$, Liangpei Zhang$^{1}$, Zhuo Zheng$^{6}$\\
$^{1}$State Key Laboratory of Information Engineering in Surveying, \\Mapping and Remote Sensing, Wuhan University, China\\
$^{2}$School of Computer Science and Technology, Xi'an Jiaotong University, China\\
$^{3}$School of Computer Science, Wuhan University, China
$^{4}$Zhongguancun Academy, China\\
$^{5}$Graduate School of Frontier Sciences, The University of Tokyo, Japan\\
$^{6}$Department of Computer Science, Stanford University, USA
}


\twocolumn[{%
\maketitle
\begin{center}
\vspace*{-1.5em}
\includegraphics[width=\textwidth]{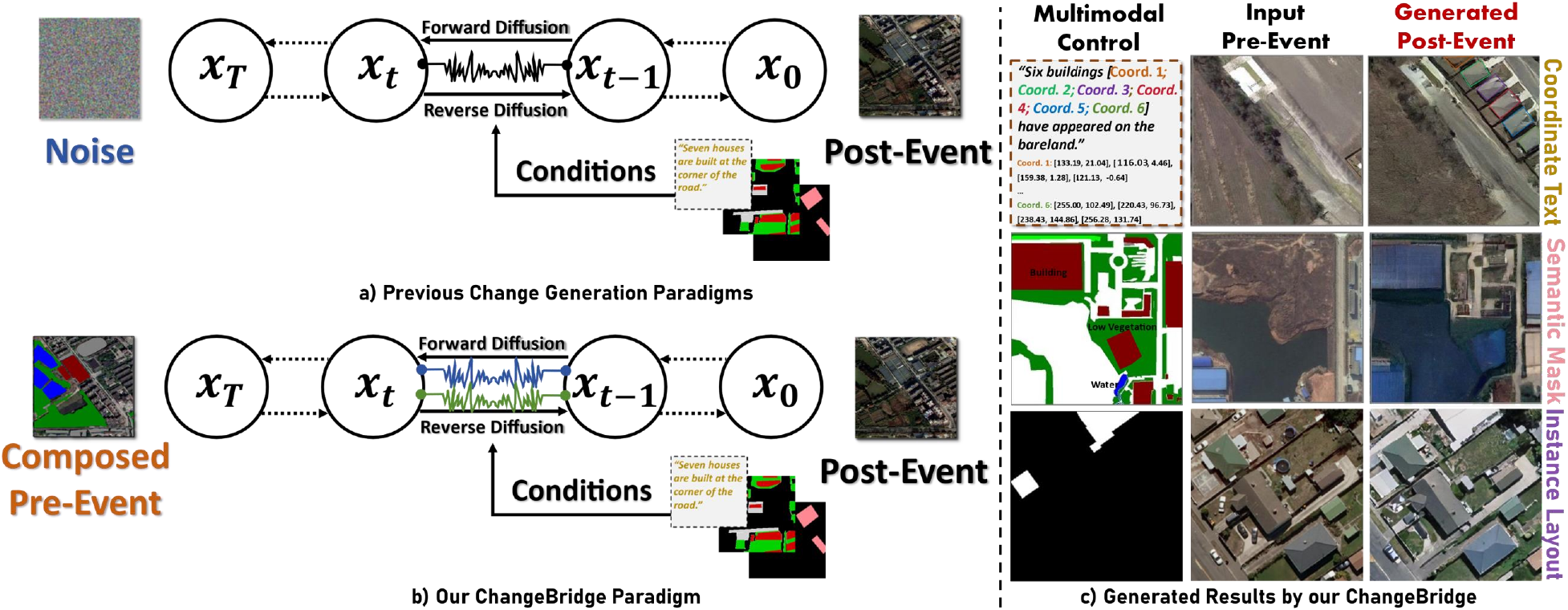}
\vspace*{-7mm}
\captionof{figure}{\textbf{Comparison of our ChangeBridge with previous change generation paradigms}. Unlike previous methods that rely on noise initialization, ChangeBridge starts from the composed pre-event state, with asynchronous drift diffusion, building a cross-spatiotemporal diffusion bridge process. It can generate post-event images based on given pre-event observations and multimodal controls, including \textbf{\textcolor{orange}{coordinate texts}, \textcolor{blue}{semantic masks}, or \textcolor{purple}{instance layouts}}. Zooming in provides better visualization.}
\label{figs:teaser}
\end{center}
}]

\renewcommand{\thefootnote}{\dag} 
\footnotetext[1]{Corresponding author: Chen Wu and Xiangyong Cao}
\renewcommand{\thefootnote}{\arabic{footnote}} 
                 
\vspace{-2mm}
\begin{abstract}
Spatiotemporal image generation is a highly meaningful task, which can generate future scenes conditioned on given observations. However, existing change generation methods can only handle event-driven changes (e.g., new buildings) and fail to model cross-temporal variations (e.g., seasonal shifts). In this work, we propose ChangeBridge, a conditional spatiotemporal image generation model for remote sensing. Given pre-event images and multimodal event controls, ChangeBridge generates post-event scenes that are both spatially and temporally coherent. The core idea is a drift-asynchronous diffusion bridge. Specifically, it consists of three main modules: a) Composed Bridge Initialization, which replaces noise initialization. It starts the diffusion from a composed pre-event state, modeling a diffusion bridge process. b) Asynchronous Drift Diffusion, which uses a pixel-wise drift map, assigning different drift magnitudes to event and temporal evolution. This enables differentiated generation during the pre-to-post transition. c) Drift-Aware Denoising, which embeds the drift map into the denoising network, guiding drift-aware reconstruction.
Experiments show that ChangeBridge can generate better cross-spatiotemporal aligned scenarios compared to state-of-the-art methods. Additionally, ChangeBridge shows great potential for land-use planning and as a data generation engine for a series of change detection tasks. Code is available at \url{https://github.com/zhenghuizhao/ChangeBridge}
\vspace{-5mm}
\end{abstract}
\section{Introduction}
Remote-sensing generation methods have been significantly advanced by the development of generative techniques. These advanced methods include layout-to-image synthesis \cite{tang2025aerogen,11180796}, modality transfer \cite{liu2023diverse,chen2024spectral}, resolution modification \cite{yu2024metaearth,changen2}, and text-to-image generation \cite{khanna2023diffusionsat,tang2024crs}. However, despite the diversity of these generative approaches, a key challenge is rarely explored: synthesizing future scenarios based on past observations and multimodal conditions, i.e., conditional spatiotemporal image generation. Addressing this challenge is crucial as it unlocks two critical capabilities. First, it provides a powerful ``what-if'' simulation tool for real-world applications like urban planning, land management, and scenario forecasting \cite{Lutjens2024Physically}. Second, it holds greater significance for the computer vision community, as it can function as a generative data engine. This can solve the severe data scarcity bottleneck for downstream spatiotemporal tasks (e.g., change detection), which require massive amounts of paired, pixel-aligned pre- and post-event data for training.

At its core, this spatiotemporal task is uniquely challenging because it must model a heterogeneous evolution. Specifically, the model must simultaneously generate: 1) drastic, event-driven changes in the foreground (e.g., new buildings emerging or regions being destroyed) and 2) subtle, temporal dynamics in the background (e.g., natural lighting shifts or vegetation growth). 

Existing change generation models~\cite{changen,zang2024ChangeDiff,changen2} follow the pipeline of Figure~\ref{figs:teaser}(a), which synthesizes changes by modifying spatial conditions, achieving the generation of event-driven changes. However, without cross-temporal modeling, they are unable to synthesize temporal dynamics. Therefore, these paradigms are ill-suited for handling this asynchronous, dual-natured generation.

To address this challenge, we propose ChangeBridge, a Drift-Asynchronous Diffusion Bridge designed to model this heterogeneous evolution. Instead of starting from pure noise, ChangeBridge directly bridges the pre-event and post-event states. Its architecture consists of three key modules: (a) composed bridge initialization, which starts the process from a composed pre-event state that coherently merges the pre-event background with the control-driven foreground; (b) asynchronous drift diffusion, which introduces a pixel-wise drift map to assign different evolution magnitudes to the foreground (high drift) and background (low drift); and (c) drift-aware denoising, where the denoising network is explicitly conditioned on this drift map to guide the differential reconstruction.

Functionally, ChangeBridge can generate realistic post-event images from a given pre-event image, allowing flexible multimodal event controls such as coordinate texts, instance layouts, or semantic masks. We evaluate two variants of ChangeBridge, utilizing three multimodal controls, on four datasets and compare their performance against six baselines. We demonstrate its effectiveness for future scenario simulation, and as a powerful data engine for downstream tasks, including change captioning and various forms of change detection. 

Our contributions can be summarized as follows:
\begin{itemize}
    \item We first introduce the task of conditional spatiotemporal image generation in remote sensing, simulating realistic future scenarios under multimodal controls.
    \item We propose ChangeBridge, a drift-asynchronous spatiotemporal diffusion bridge generative model, enabling conditions on instance layouts, semantic masks, and coordinate texts.
    \item We demonstrate that ChangeBridge can serve as a powerful generative data engine, significantly improving downstream change detection performance. 

\end{itemize}

\section{Related work}
\label{sec:related_work}
\subsection{Spatiotemporal Scenario Simulation}
Spatiotemporal scenario simulation plays a crucial role in understanding land use changes and urban dynamics in remote sensing. Traditional methods in this area primarily rely on rule-based algorithms \cite{roodposhti2019novel, chaturvedi2021machine, crooks2010rule} and statistical models \cite{ren2019spatially, tong2022statistical} to perform numerical simulations of spatiotemporal dynamics. Still, they fail to synthesize realistic future scenarios that are visually intuitive and easy to interpret.

Recently, generative models have revitalized scene-change generation. Several studies use GAN-based and diffusion-based frameworks to synthesize changes by modifying spatial conditions like maps, masks, or text-driven layouts \cite{changen,changen2,changeanywhere,zang2024ChangeDiff}. These methods focus on spatial differences, and cannot model cross-temporal transitions from historical observations. Lütjens et al. \cite{Lutjens2024Physically} made an initial attempt at spatiotemporal generation, using a GAN-based~\cite{wang2018pix2pixHD} framework for climate visualization, supporting the instance-layout condition. However, it can only simulate disaster weather from the current scene and falls short in terms of precise conditional control and universality. In this work, we first explore the general paradigm of spatiotemporal image generation with multimodal controls.

\begin{figure*}
  \vspace{-5em}
	\centering
	\includegraphics[width=1.0\linewidth
]{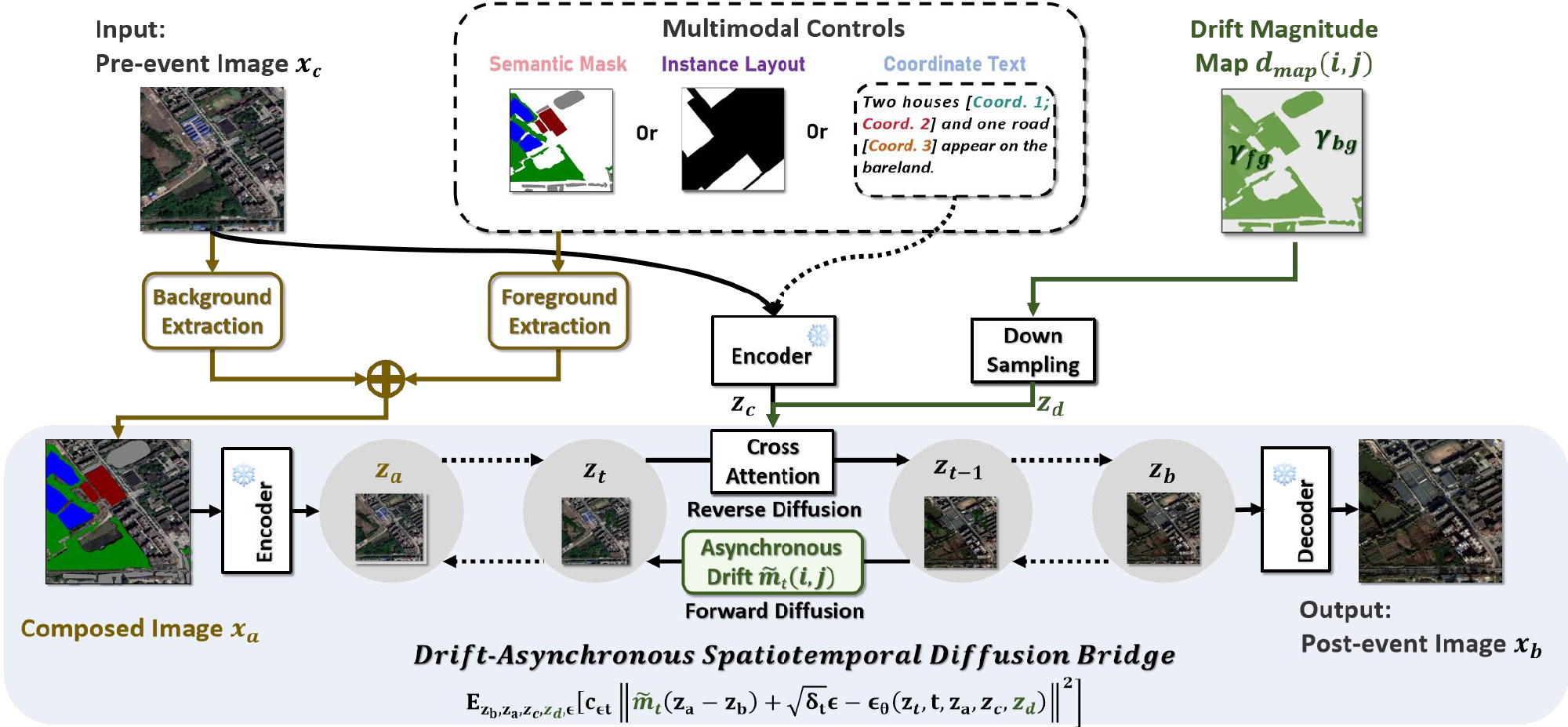}
	\vspace{-1.5em}
    \caption{\textbf{Framework of ChangeBridge.} It includes three main modules: a) \textbf{Composed bridge initialization:} builds a cross-spatiotemporal diffusion bridge from a composed input, combining the pre-event background with the control-driven foreground. b) \textbf{Asynchronous drift diffusion:} introduces a pixel-wise drift map to assign different drift magnitudes to the foreground and background regions, enabling differentiated evolution. c) \textbf{Drift-aware denoising:} embeds the drift distribution into the denoising network to guide drift-aware reconstruction. Multimodal controls (e.g., coordinate text, semantic mask, and instance layout) are supported.}
	\label{method1}

\end{figure*}

\subsection{Multi-Conditional Generation}
Multi-condition models, especially diffusion-based ones, are widely used for controllable generation.
These models condition on spatial layouts \cite{li2023gligen}, edges \cite{zhang2023inversion}, depth maps \cite{zhang2023inversion}, style exemplars \cite{sasaki2021unit,Mo_2024_CVPR}, and textual prompts \cite{ruiz2023dreambooth}. They have also been extended to hierarchical layout control \cite{DBLP:conf/nips/ChengMwLMWLY24}, multi-object image generation \cite{ijcai2025p217}, and scientific structure design under complex property constraints \cite{park2025moffusion}.

Although these approaches achieve strong controllability, they still follow a noise-to-image diffusion process.
During denoising, they usually rely on additional techniques, such as classifier-free or guidance-free sampling \cite{dhariwal2021diffusion}, LoRA adaptation \cite{yang2024lora}, or projected sampling \cite{yu2023video}, to align multiple conditions. Since generation starts from pure noise, the noise-to-image model must reconstruct both content and structure from scratch. It is difficult to maintain cross-image spatial coherence and temporal consistency. This limitation becomes especially evident in spatiotemporal generation, where transitions should remain consistent with historical observations under multiple controls.

\subsection{Diffusion Bridge Model}
\vspace{-1mm}
The diffusion bridge formulation replaces the conventional “noise-to-state” generation paradigm with a “state-to-state” transformation, enabling explicit modeling of structured transitions between paired observations.

Recent studies have explored various bridge-based formulations, including Brownian Bridge diffusion models for image-to-image translation \cite{li2023bbdm}, denoising diffusion bridge models for stochastic distribution transport \cite{zhou2023denoising}, and Schrödinger-bridge frameworks that generalize diffusion models for bidirectional distribution matching \cite{de2021diffusion,shi2023dsbm}.
Unlike conventional diffusion models that start from noise, bridge-based methods establish explicit mappings between two structured states or distributions, naturally preserving structural consistency and semantic alignment throughout the generation process \cite{li2023bbdm,zhou2023denoising,shi2023dsbm}. Inspired by these advances and the need to preserve background structure in remote sensing, we propose a diffusion-bridge model for spatiotemporal generation with multimodal controls, enabling stable transitions between pre- and post-event images.

\section{Methodology}
\vspace{-1mm}
We propose ChangeBridge, an asynchronous spatiotemporal diffusion bridge. 
ChangeBridge consists of three main components: a) Composed Bridge Initialization, as detailed in Section~\ref{sec_b}; b) Asynchronous Drift Modeling, as described in Section~\ref{sec_c}; and c) Drift-Aware Reverse Denoising, as presented in Section~\ref{sec_d}. The overall ChangeBridge framework is illustrated in Figure~\ref{method1}.


\subsection{Preliminary}
We first introduce the definition of diffusion bridge models, which extend the conventional diffusion generation from noise-based reconstruction to structured state transitions between two endpoints~\cite{li2023bbdm}. 
We adopt the classical Brownian bridge formulation, short for \emph{diffusion bridge}. 

Given two related images \( \bm{x}_a \) and \( \bm{x}_b \), a pre-trained autoencoder \( \mathcal{E} \) encodes them into latent representations to reduce computational cost, i.e., \( \bm{z}_a = \mathcal{E}(\bm{x}_a) \) and \( \bm{z}_b = \mathcal{E}(\bm{x}_b) \), where \( \bm{z}_a, \bm{z}_b \sim q(\bm{z}) \). Following the diffusion bridge process~\cite{wang1945theory,pollak1985diffusion,li2023bbdm}, 
the latent state at timestep \( t \in [0, T] \), evolving from \( \bm{z}_0 \) to \( \bm{z}_T \), can be expressed as:
\begin{equation}\small
\label{eq_bbp}
p(\bm{z}_t | \bm{z}_0, \bm{z}_T) 
= \mathcal{N}\!\left( (1 - \frac{t}{T}) \bm{z}_0 + \frac{t}{T} \bm{z}_T,\; \frac{t(T-t)}{T} \bm{I} \right).
\end{equation}

\noindent\textbf{Forward process.} The diffusion bridge constructs a stochastic trajectory that connects two structured endpoints \( \bm{z}_b \to \bm{z}_a \) over \( T \) timesteps. 
Each intermediate latent \( \bm{z}_t \) follows a Gaussian distribution:
\begin{equation}\small
\label{eq_fp}
q(\bm{z}_t \mid \bm{z}_b, \bm{z}_a)
= \mathcal{N}\!\big(\bm{z}_t;\, (1-m_t)\bm{z}_b + m_t\bm{z}_a,\, \delta_t \bm{I}\big),
\end{equation}
where \( m_t = \tfrac{t}{T} \) denotes the drift term and \( \delta_t = 2(m_t - m_t^2) \) represents the variance term along the bridge. This process defines a Markov chain. Unlike the standard diffusion model, which is a drift-free case of pure noise injection, the diffusion bridge introduces a non-zero drift \( m_t \). This drift \( m_t\) progressively connects two structured endpoints, enabling a directional transition across timesteps. 

\noindent\textbf{Reverse process.} The reverse process reconstructs the latent state \( \bm{z}_b \) starting from \( \bm{z}_a \) through a parameterized denoising network \( \bm{\epsilon}_{\theta} \), modeled as:
\begin{equation}\small
\label{eq_bp}
p_{\theta}(\bm{z}_{t-1} | \bm{z}_t, \bm{z}_a)
= \mathcal{N}\!\big(\bm{z}_{t-1};\, \hat{\bm{\mu}}_{\theta}(\bm{z}_t, t, \bm{z}_a),\, \hat{\sigma}^2_{\theta}(t) \bm{I}\big).
\end{equation}
This reverse process learns the stochastic evolution from the latent state \( \bm{z}_a \) to \( \bm{z}_b \), directly modeling cross-state dependencies rather than learning from a random prior.

\noindent\textbf{Training objective.} The denoising network is trained to predict the perturbation \( \bm{\epsilon}\) along the bridge trajectory:
\begin{equation}\small
\label{eq_loss}
\mathcal{L} = 
\mathbb{E}_{\bm{z}_a, \bm{z}_b, \bm{\epsilon}}\!
\left[\!\big\| m_t(\bm{z}_a - \bm{z}_b) 
+ \sqrt{\delta_t}\bm{\epsilon}
- \bm{\epsilon}_{\theta}(\bm{z}_t, t, \bm{z}_a)\big\|^2\!\right].
\end{equation}
Finally, the reconstructed latent \( \bm{z}'_b \) is decoded by the pre-trained autodecoder \( \mathcal{D} \) to generate the image \( \bm{x}'_b = \mathcal{D}(\bm{z}'_b) \).

\subsection{Composed Bridge Initialization} \label{sec_b}
\begin{figure}
	\centering
	\includegraphics[width=1\linewidth
]{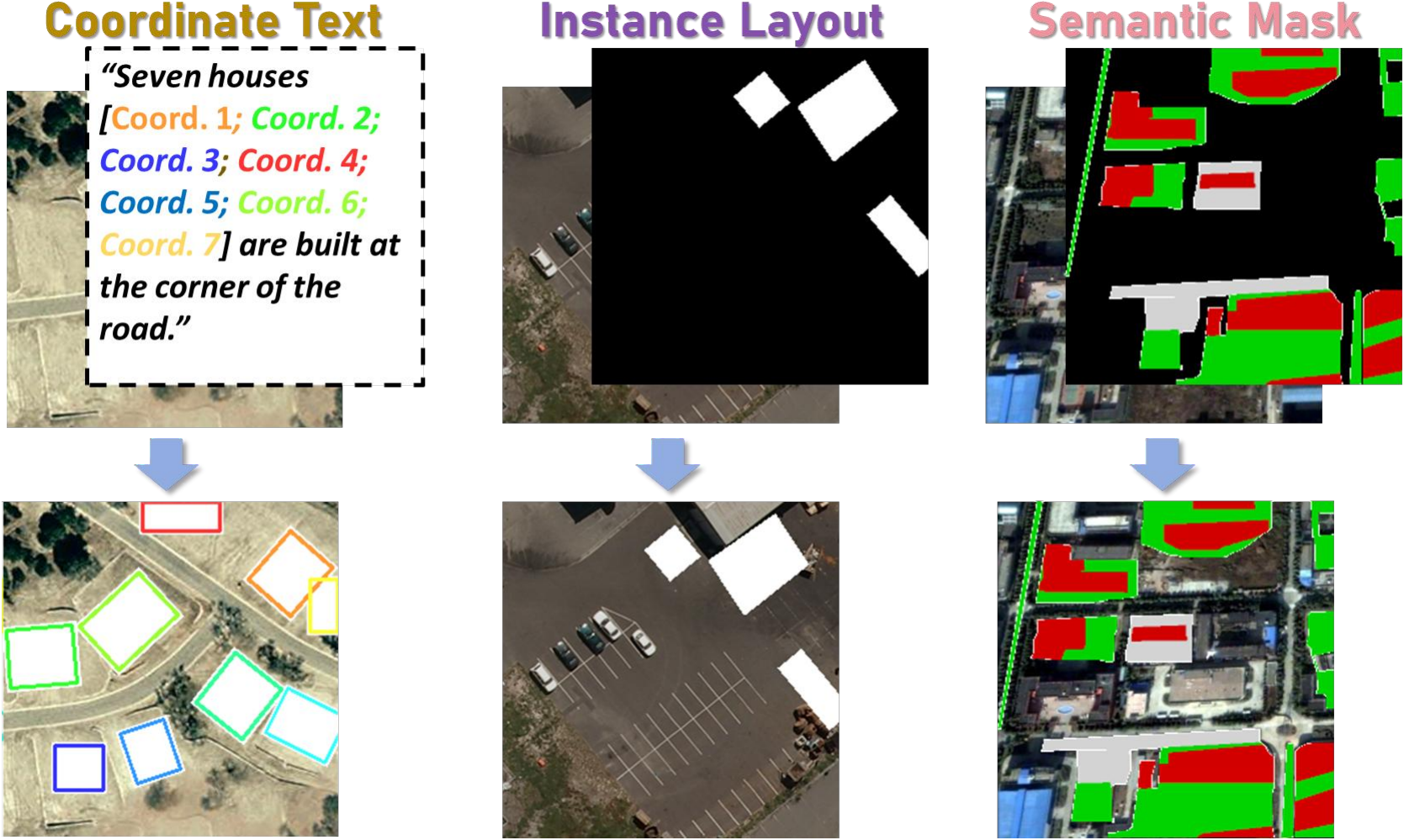}
	\caption{\textbf{Examples of composed images}. Each image combines the background from the pre-event image with the control-localized foreground.}
	\label{method2}
    \vspace{-1em}
\end{figure}

To enable spatiotemporal image generation, we construct composed images as the initialization of the diffusion bridge. This composition integrates the pre-event background and multimodal condition-guided foreground, providing context priors to jointly model foreground events and background temporal evolution. The composed image inputs are shown in Figure ~\ref{method2}.

\noindent\textbf{Conditional spatial awareness.} Given a multimodal condition \( \bm{x}_c \), which can be one of instance layouts, semantic masks, or coordinate texts. We localize event-relevant regions \( \bm{M}_{\text{fg}} \in [0,1]^{1\times H\times W} \), and obtain a binary spatial prior according to the selected condition as follows:

\begin{itemize}[leftmargin=1.2em]
    \item \textbf{Instance layout and semantic mask.}
Each condition image maps event and no-event regions through color channels. We assume that a specific color value \( \bm{c}_{\text{bg}} \) represents the no-event regions, i.e., background.
    The binary mask is then derived by pixel-wise comparison $ \small
    \bm{M}_{\text{fg}}(i,j) =
    \begin{cases}
    0, & \text{if } \bm{x}_c(i,j) = \bm{c}_{\text{bg}}\\
    1, & \text{otherwise}
    \end{cases}$. This definition ensures that all non-background event regions are explicitly highlighted as foreground.

    \item \textbf{Coordinate text.} Each coordinate-text pair, such as ``Six buildings [Coords. 1–6] appear on the bareland,'' provides six spatially-localized coordinates of six buildings. For the \( k \)-th object in $N_o$ coordinates, four corner points are given $\bm{P}_k = \{(x_1, y_1), (x_2, y_1), (x_1, y_2), (x_2, y_2)\}$, and $ k = 1,\dots,N_o$. Each \( \bm{P}_k \) represents the Cartesian product of two horizontal and two vertical coordinates, which can form the four vertices of a rotated bounding box in the image plane. So we fit a rotated bounding box $\small
    \mathcal{B}_k = \operatorname{RotBox}(\bm{P}_k) 
    = \{(x, y) \mid \bm{R}(\theta_k)\bm{p} + \bm{t}_k \in \Omega_k\}$, where \( \bm{R}(\theta_k) \) denotes a 2D rotation matrix with angle \( \theta_k \), \( \bm{t}_k \) is the box center, and \( \Omega_k \subset \mathbb{R}^2 \) defines the local rectangular support. The foreground mask induced by all described objects is similarly defined as $\small
    \bm{M}_{\text{fg}}(i,j) =
    \begin{cases}
    1, & \text{if } (i,j) \in \bigcup_k \mathcal{B}_k\\
    0, & \text{otherwise}
    \end{cases}
$,
This mapping \( f_{\text{text}} : (\bm{P}_k, t_k) \mapsto \bm{M}_{\text{fg}}^{\text{text}} \) transforms coordinate-text pairs into a binary spatial mask, indicating event-relevant regions described by natural language.
\end{itemize}

\noindent\textbf{Foreground–background extraction.} According to the condition-aware foreground mask \( \bm{M}_{\text{fg}} \in [0,1]^{1 \times H \times W} \), we define the complementary background mask as \( \bm{M}_{\text{bg}} = 1 - \bm{M}_{\text{fg}} \).  
The background regions are extracted from the pre-event image \( \bm{x}_0 \) via element-wise multiplication \( \bm{x}_{\text{bg}} = \bm{M}_{\text{bg}} \odot \bm{x}_0 \). The corresponding foreground regions  $\bm{x}_{\text{fg}}$ are preserved from the condition \( \bm{x}_c \) that \( \bm{x}_{\text{fg}} = \bm{M}_{\text{fg}} \odot \bm{x}_c \),  
where \( \odot \) denotes element-wise multiplication.  

\noindent\textbf{Composed image input.} By combining the condition-driven foreground and pre-event background,  
we obtain the composed initial state of the diffusion bridge \( \bm{x}_a = \bm{x}_{\text{fg}} + \bm{x}_{\text{bg}} \).  
This composition serves as the initial state of the diffusion bridge.

\subsection{Asynchronous Drift Diffusion}  \label{sec_c}
Formally, let \( \bm{x}_a \) denote the composed bridge input introduced above, obtained by merging the condition-driven foreground with the pre-event background. Let \( \bm{x}_b \) denote the post-event remote sensing image.  
Their latent representations \( \bm{z}_a = \mathcal{E}(\bm{x}_a) \) and \( \bm{z}_b = \mathcal{E}(\bm{x}_b) \) are obtained through an autoencoder \( \mathcal{E} \), where \( \bm{z}_a, \bm{z}_b \in \mathbb{R}^{n \times m} \).  
We assume that the source and target latent variables follow two Gaussian distributions,  
\( \bm{z}_a \sim q_{\text{pre}}(\bm{z}_a) \) and \( \bm{z}_b \sim q_{\text{post}}(\bm{z}_b) \), both approximated by \( \mathcal{N}(\mathbf{0}, \bm{I}) \).  
Similarly, the multimodal condition \( \bm{x}_c \) is encoded into the same latent space through a domain-specific encoder \( \tau_{\theta} \), i.e., \( \bm{z}_c = \tau_{\theta}(\bm{x}_c) \in \mathbb{R}^{n \times m} \). 

The drift-asynchronous diffusion process models the spatially varying drift between the composed bridge latent \( \bm{z}_a \) and the post-event target latent \( \bm{z}_b \), which is specifically described in detail as follows.

\noindent\textbf{Drift magnitude map.} Given the binary foreground mask \( \bm{M}_{\text{fg}} \in \{0,1\}^{ H \times W} \) derived from multimodal conditions, we construct a pixel-wise drift magnitude map \( \bm{d}_{\text{map}} \) to control the local evolution intensity during the diffusion process:
\begin{equation}\small
\bm{d}_{\text{map}}(i,j)
= \bm{M}_{\text{fg}} \cdot \gamma^{fg}
+ \big(1 - \bm{M}_{\text{fg}}\big) \cdot \gamma^{bg},
\label{eq:dmap}
\end{equation}
where \( \gamma^{fg} \) and \( \gamma^{bg} \) denote the drift magnitudes for the foreground and background evolutions, respectively, with \( \gamma^{fg} > \gamma^{bg} \) by design. Then, the resulting map \( \bm{d}_{\text{map}} \) is downsampled by adaptive average pooling to obtain the latent of the drift magnitude map \( \bm{z}_d \), and is further normalized to \([0,1]\) for numerical stability. This drift map latent \( \bm{z}_d \) is spatially aligned with the latent variables \( \bm{z}_a \) and \( \bm{z}_b \), providing consistent modulation across regions.

\noindent\textbf{Forward process.} Building upon the diffusion bridge formulation in Eq.~(\ref{eq_fp}), we introduce a spatially asynchronous drift map \( \bm{d}_{\text{map}} \) that modulates the drift magnitude across different spatial regions. 

Specifically, for each pixel location \( (i,j) \), we redefine the drift coefficient as 
\(\tilde{m}_t(i,j) = m_t \cdot {\bm{z}_d}(i,j)\), 
where \(m_t = \tfrac{t}{T}\) is the canonical bridge coefficient defined in Eq.~(\ref{eq_fp}). This allows spatially adaptive drift velocities for region-specific spatiotemporal transitions.
The forward transition can then be defined as:
\begin{equation}\small
\label{eq:asy_fp}
q(\bm{z}_t \mid \bm{z}_b, \bm{z}_a)
= \mathcal{N}\!\big(
\bm{z}_t;\,
(1-\tilde{m}_t(i,j))\bm{z}_b + \tilde{m}_t(i,j)\bm{z}_a,\,
\delta_t \bm{I}
\big),
\end{equation}
where \( \delta_t = 2(m_t - m_t^2) \) is the variance term inherited from the original definition in Eq.~(\ref{eq_fp}).  It is an empirically validated design to keep the variance schedule \( \delta_t \) consistent with the original diffusion bridge for training stability. This design can be viewed as a drift-generalized version of the Brownian bridge diffusion~\cite{li2023bbdm}, and the theoretically consistent counterpart with the asynchronous variance \( \tilde{\delta}_t \) is provided in the supplementary.

\subsection{Drift-Aware Denoising} \label{sec_d}
During the reverse process, we further incorporate the spatially varying drift latent \( \bm{z}_d \) to adaptively reconstruct the heterogeneous foreground and background dynamics. 

\noindent\textbf{Reverse process.} Following the reverse dynamics in Eq.~(\ref{eq_bp}), the model reconstructs the target latent state \( \bm{z}_b \) from the composed latent state \( \bm{z}_a \) under the guidance of the spatially varying drift. The reverse transition is as follows:
\begin{equation}\small
\label{eq:asy_bp}
p_{\theta}(\bm{z}_{t-1} | \bm{z}_t, \bm{z}_a, \bm{z}_c)
= \mathcal{N}\!\big(
\bm{z}_{t-1};\,
\hat{\bm{\mu}}_{\theta}(\bm{z}_t,t, \bm{z}_a, \bm{z}_c, \bm{z}_d),\,
\hat{\sigma}^2_{\theta}(t) \bm{I}
\big),
\end{equation}
where the drift term is modulated by the spatial magnitude map \( \bm{z}_d \), allowing region-dependen, directional guidance during denoising.  Here \( \bm{z}_c \) denotes the pre-event image latent, providing the global context of the current observation scene. For the coordinate–text condition, we further concatenate the pre-event latent and the coordinate–text latent as \( \bm{z}_c = [\,\bm{z}_{\text{text}} \,\|\, \bm{z}_{\text{pre}}\,] \), thereby integrating the pre-event context with event-relevant semantics.

\noindent\textbf{Training objective.} To learn the asynchronous dynamics, the denoising network \( \bm{\epsilon}_{\theta} \) is trained to predict the perturbation noise $ \bm{\epsilon}$ under the drift-modulated forward process. The loss is defined based on Eq.~(\ref{eq_loss}) with pixel-wise drift magnitude maps:
\begin{equation}\small
\label{eq:asy_loss}
\mathcal{L}_{\text{asy}}
=
\mathbb{E}_{\bm{z}_a, \bm{z}_b, \bm{z}_c, \bm{\epsilon}}\!
\Big[
\big\|
\tilde{m}_t(\bm{z}_a - \bm{z}_b)
+ \sqrt{\delta_t}\bm{\epsilon} - \bm{\epsilon}_{\theta}(\bm{z}_t,t, \bm{z}_a, \bm{z}_c, \bm{z}_d)
\big\|^2
\Big]
\end{equation}
This objective modifies the drift term from Eq.~\ref{eq_loss}, introducing a spatially adaptive drift coefficient while preserving the same variance schedule \( \delta_t \). As a result, the model learns heterogeneous transition behaviors, generating stronger evolution in event-related regions, and simulating smoother temporal dynamics in background areas.

\begin{figure*}
	\centering
      \vspace{-3em}
	\includegraphics[width=\linewidth]{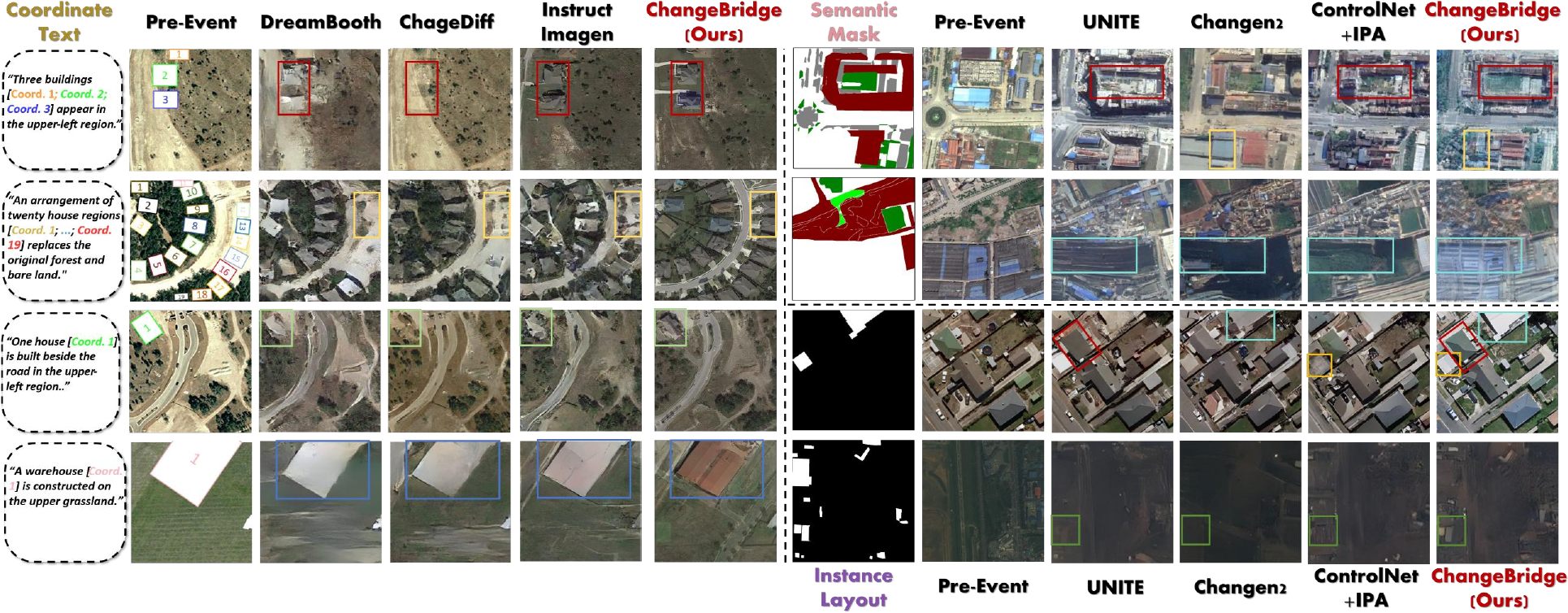}
	\vspace{-2em}
    \caption{\textbf{Qualitative comparison} conditioned on \textbf{\textcolor{orange}{coordinate texts}, \textcolor{blue}{semantic masks}, and \textcolor{purple}{ instance layouts}}. Colored boxes highlight corresponding regions across methods (same color = same region; different colors = different regions). Zooming in provides better visualization. }
	\label{experiment1}
    \vspace{-1em}
\end{figure*}
\vspace{-1mm}
\subsection{Practical Implementation} \label{sec:e}
\vspace{-1mm}
\noindent\textbf{Pipeline overview.} This formulation is backbone-independent, making it applicable to both convolutional and transformer-based diffusion architectures. The pipeline follows as:
1) Given the \textit{latent variable \( \bm{z}_t \)}, which evolves from the \emph{composed image latent} \( \bm{z}_a \), the model extracts multi-scale features and is guided by two additional signals during denoising: 2) the \emph{pre-event latent} \( \bm{z}_c \), providing global context information (for coordinate-text inputs, \( \bm{z}_c \) additionally includes the text latent), and 3) the \emph{drift latent} \( \bm{z}_d \), indicating how strongly the foreground event and background temporal evolution occurs. 4) The \emph{post-event latent} \( \bm{z}_b \) is jointly guided by these signals to synthesize through the reverse denoising process.

\noindent\textbf{Fusion mechanisms.} We adopt standard condition-fusion strategies to integrate the drift latent $\bm{z}_d$ with the pre-event latent $\bm{z}_c$, depending on the network backbone. 

For convolutional variants (e.g., UNet~\cite{ronneberger2015u}), we perform conditional fusion via channel-wise concatenation of $\bm{z}_c$ and $\bm{z}_d$, forming a fused representation $\text{Concat}(\bm{z}_c, \bm{z}_d)$. The fused features are processed by convolutional layers, followed by a self-attention block to capture global context. The attention operation follows the standard formulation $
\text{Attn}(\bm{Q}, \bm{K}, \bm{V}) =
\text{Softmax}\!\left(\frac{\bm{Q}\bm{K}^\top}{\sqrt{d}}\right)\bm{V}$. Here, $\bm{K}$ corresponds to the current latent feature, and $[\bm{Q}, \bm{V}]$ are derived from the fused representation. When coordinate–text conditions are available, cross-attention is further applied to align visual features with textual embeddings.

For transformer-based variants (e.g., DiT~\cite{Peebles2023DiT}), we follow its multimodal extension and adopt a FiLM-style additive modulation~\cite{perez2018film}. Specifically, the condition latents are concatenated and projected by a lightweight MLP $\phi(\cdot)$, and then added to the intermediate feature as a residual modulation: $\bm{x}'_t = \bm{x}_t + \phi(\text{Concat}(\bm{z}_c, \bm{z}_d)).$ This design enables flexible condition injection without modifying the transformer architecture.

\section{ Experiments}
\label{sec:experiments}
\subsection{Experimental Setting}
\noindent\textbf{Dataset descriptions.} We conduct experiments on four change detection benchmarks with default splits: 1) LEVIR-CC \cite{chen2020spatial}, for coordinate texts; 2) WHU-CD \cite{ji2018fully} and S2Looking \cite{shen2021s2looking}, for instance layouts; and 3) SECOND \cite{yang2020semantic}, for semantic masks. We create a coordinate-augmented version of LEVIR-CC, adding object counting and precise object-level coordinates. 

\noindent \textbf{Implementation details.} ChangeBridge uses both UNet- and DiT-based denoising backbones. The UNet variant is based on Stable Diffusion (SD)~1.5~\cite{rombach2022ldm}, while the DiT variant uses DiT-XL/2~\cite{Peebles2023DiT}, initialized from pretrained weights and modulated with FiLM-style layers following SD3.5~\cite{stabilityai2024sd35}. Both variants are trained with the Brownian-Bridge objective, using 1,000 timesteps for training and 200 for inference. The UNet and DiT variants are trained for 60 and 100 epochs, taking approximately 12 and 25 hours, respectively, on $256 \times256$ satellite images using Adam ($1.0\!\times\!10^{-4}$) with a batch size of 64 on two NVIDIA A100 GPUs. The image encoder is a pretrained VQGAN~\cite{vagan}, and the text encoder is SkyCLIP~\cite{wang2024skyscript}, fine-tuned for remote sensing. The maximum bridge variance is set to 1.0, optimized with $L_1$ loss. Asymmetric drift magnitudes are set to $\gamma^{fg}=1.0$ and $\gamma^{bg}=0.8$ for the WHU, S2Looking, and LEVIR-CC datasets, and $\gamma^{fg}=1.0$ and $\gamma^{bg}=0.7$ for SECOND. Unless specified, results use the DiT-based ChangeBridge, and ablations are conducted with the UNet variant.

\begin{table*}
\small
\centering
\caption{\textbf{Quantitative comparison} conditioned on instance layout, semantic mask, and text prompts. FID and IS are reported for all settings; mIoU (\%) for semantic masks, IoU (\%) for instance layouts, and cosine similarity (CosSim) for coordinate texts. Results on real data serve as the upper bound. ControlNet* denotes ControlNet+IPA, and Instruct* refers to Instruct-Imagen. Ours-U is our UNet variant, and Ours-T is the DiT variant.}
\renewcommand{\arraystretch}{1.2} 
\setlength{\tabcolsep}{1.5pt}     
\vspace{-0.5em}
\begin{minipage}{0.25\textwidth}  
\raggedright  
\begin{tabular}{c|ccc}
\multirow{2}{*}{\textcolor{orange}{\textbf{Text}}}& \multicolumn{3}{c}{\textbf{LEVIR-CC}}\\
& \textbf{FID} $\downarrow$& \textbf{IS} $\uparrow$&\textbf{CosSim} $\uparrow$\\ \hline
Real Data& -&  -&0.89\\ \hline
 DreamBooth& 95.64&1.52                    &0.76\\
 ChangeDiff& 55.60& 3.43&0.79\\
Instruct*& 48.17& 3.70&0.81\\
\hline
\textbf{Ours-U}
& \textcolor{orange}{\textbf{38.36}}& \textcolor{orange}{\textbf{4.01}}&\textcolor{orange}{\textbf{0.82}}\\
 \textbf{Ours-T}& \textcolor{orange}{\textbf{31.45}}& \textcolor{orange}{\textbf{5.14}}&\textcolor{orange}{\textbf{0.85}}\\ 
\end{tabular}
\end{minipage}
\hfill
\begin{minipage}{0.42\textwidth}  
\centering  
\begin{tabular}{c|ccc|ccc}
\multirow{2}{*}{\textcolor{purple}{\textbf{Layout}}}& \multicolumn{3}{c}{\textbf{WHU-CD}}&\multicolumn{3}{c}{\textbf{S2Looking}}\\
& \textbf{FID} $\downarrow$  &\textbf{IS} $\uparrow$  
& \textbf{IoU} $\uparrow$&\textbf{FID} $\downarrow$ & \textbf{IS} $\uparrow$ 
& \textbf{IoU} $\uparrow$\\ \hline
Real Data&  -&
-&  81.30&-&    
-& 82.47\\ \hline
 UNITE                 & 86.81&5.76                    
& 69.38&97.57& 3.98                   
&72.61\\
ControlNet*& 52.08                      &5.12                    
& 71.54&94.68                     & 4.56                   
& 75.20\\
 Changen2& 48.85& 5.64& 74.33& 83.31& 4.02&78.89\\
\hline
\textbf{Ours-U}& \textcolor{purple}{\textbf{45.47}}&\textcolor{purple}{\textbf{5.88}}& \textcolor{purple}{\textbf{75.30}}&\textcolor{purple}{\textbf{72.56}}& \textcolor{purple}{\textbf{4.60}}& \textcolor{purple}{\textbf{78.45}}\\
 \textbf{Ours-T}& \textcolor{purple}{\textbf{40.12}}& \textcolor{purple}{\textbf{6.77}}& \textcolor{purple}{\textbf{78.13}}& \textcolor{purple}{\textbf{56.42}}& \textcolor{purple}{\textbf{5.22}}&\textcolor{purple}{\textbf{79.40}}\\ 
\end{tabular}
\end{minipage}
\hfill
\begin{minipage}{0.25\textwidth}  
\raggedleft  
\begin{tabular}{c|ccc}
\multirow{2}{*}{\textcolor{blue}{\textbf{Semantic}}}& \multicolumn{3}{c}{\textbf{SECOND}}\\
& \textbf{FID} $\downarrow$ & \textbf{IS} $\uparrow$  &\textbf{mIoU} $\uparrow$\\ \hline
Real Data& -&  -&76.19\\ \hline
 UNITE                 & 78.42&5.72&70.22\\
ControlNet*& 90.81& 3.31 &72.30\\
 Changen2& 69.43& 6.18&73.20\\
\hline
\textbf{Ours-U}& \textcolor{blue}{\textbf{62.24}}& \textcolor{blue}{\textbf{6.03}}&\textcolor{blue}{\textbf{73.47}}\\
 \textbf{Ours-T}& \textcolor{blue}{\textbf{59.33}}& \textcolor{blue}{\textbf{6.41}}&\textcolor{blue}{\textbf{74.26}}\\ 
\end{tabular}
\end{minipage}
\vspace{-0.5em}
\label{tab_1}
\end{table*}

\subsection{Comparison with Prior Methods}
\noindent \textbf{Baseline models.} We compare our ChangeBridge method against six existing methods: four multi-conditional generation methods (DreamBooth \cite{ruiz2023dreambooth}, Instruct-Imagen \cite{hu2024instruct}, UNITE \cite{zhan2021unbalanced}, ControlNet \cite{zhang2023adding} with IP-Adapter) \cite{ye2023ipadapter}, and two change generation methods (Changen2~\cite{changen2} and ChangeDiff~\cite{zang2024ChangeDiff}). For multi-conditional methods, the pre-event image serves as the source, multimodal controls as the condition, and the post-event image as the target. For change generation methods, we use their original pipeline, inputting only pre-event images and controls.

\noindent \textbf{Qualitative results.} Figure~\ref{experiment1} presents the qualitative comparisons of ChangeBridge, divided into three aspects:
\textit{1) Coordinate Text:} As shown on the left of Figure~\ref{experiment1}, DreamBooth often misaligns the generated buildings with the given coordinates, while Instruct-Imagen lacks fine-grained details. In contrast, ChangeBridge precisely follows the coordinate texts, preserves the scene structure, and produces clearer images.
\textit{2) Instance Layout:} As shown on the bottom-right of Figure~\ref{experiment1}, ChangeBridge aligns better with the provided spatial layouts than the other baseline methods. Compared to the change generation methods, it also produces more coherent cross-spatiotemporal evolution (e.g., lighting shifts) in the first example.
\textit{3) Semantic Mask:} As shown in the upper right of Figure~\ref{experiment1}, ChangeBridge demonstrates a clear improvement in semantic alignment over the three baseline methods across many regions. It also follows the instance layout constraints more accurately than the other three competing methods (Changen2, ControlNet+IPA, and UNITE), and preserves background consistency better.

\begin{figure}
	\centering
	\includegraphics[width=1\linewidth]{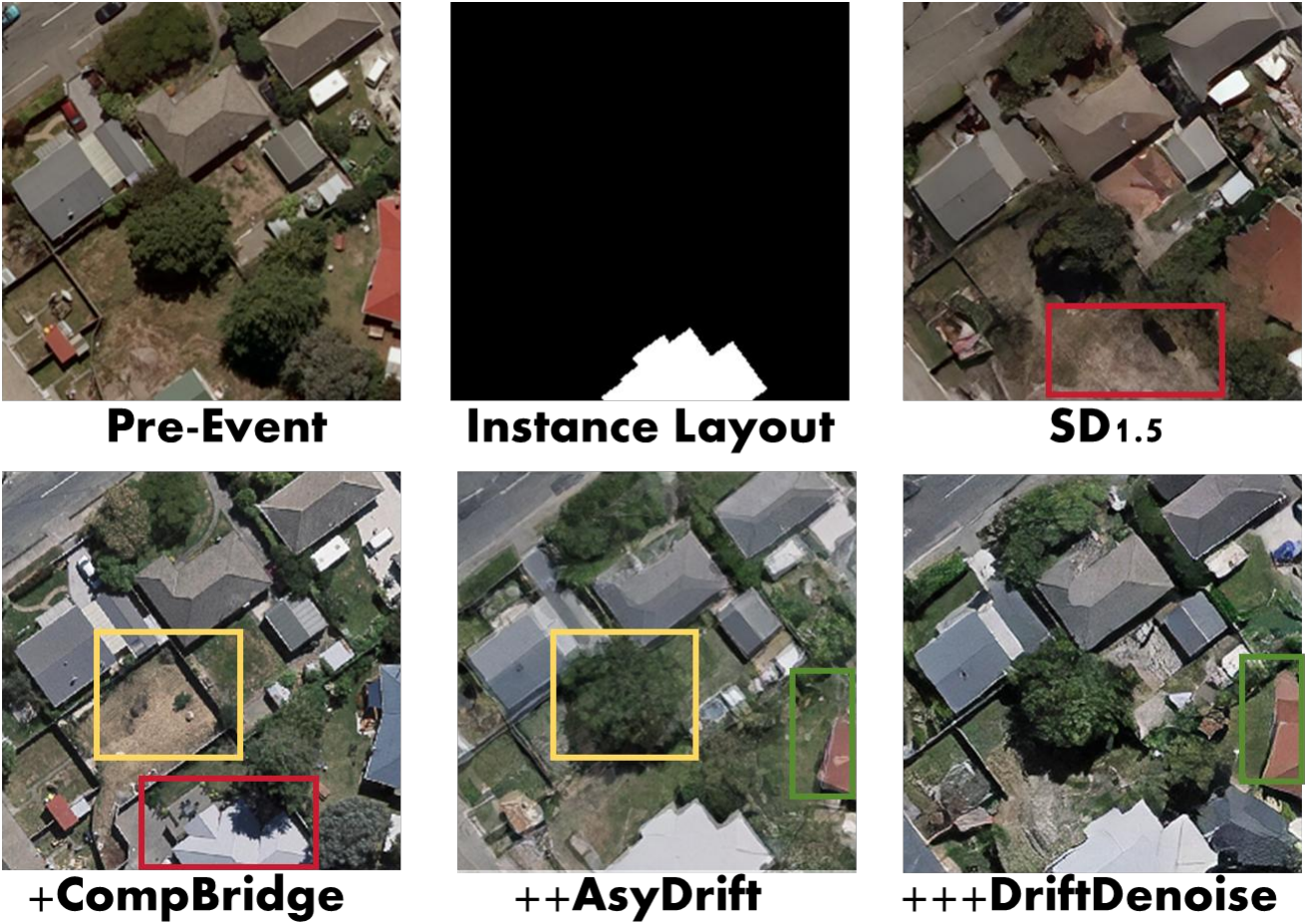}
	\caption{\textbf{Qualitative ablation} of ChangeBridge components. Starting from the SD1.5 baseline, we progressively incorporate CompBridge, AsyDrift, and DriftDenoise. Colored boxes highlight corresponding regions across methods (same color = same region; different colors = different regions).}
	\label{experiment3-1}
\end{figure}

\begin{table}
\setlength{\tabcolsep}{6pt}
\renewcommand{\arraystretch}{1.2}
\centering
\caption{\textbf{Quantitative ablation} of ChangeBridge components. FID, IS, and IoU (\%) are reported on WHU. CB denotes the CompBridge module, AD denotes AsyDrift, and DD denotes DriftDenoise.}
\begin{tabular}{cccccc} 
\hline\hline
\textbf{CB} & \textbf{AD} & \textbf{DD} & \textbf{FID $\downarrow$} & \textbf{IS $\uparrow$} & \textbf{IoU $\uparrow$} \\ 
\hline
& & & 76.81 & 4.85 & 65.29 \\
\textbf{\textcolor{blue!50}{\checkmark}}& & & 56.24 \tiny\textcolor{red}{-20.57}& 5.43 \tiny\textcolor{red}{+0.58} & 71.87 \tiny\textcolor{red}{+6.58} \\
\textbf{\textcolor{blue!50}{\checkmark}}& \textbf{\textcolor{olive}{\checkmark}}& & 57.06 \tiny\textcolor{blue}{+0.82}& 5.61 \tiny\textcolor{red}{+0.18} & 72.48 \tiny\textcolor{red}{+0.61} \\
\textbf{\textcolor{blue!50}{\checkmark}}& \textbf{\textcolor{olive}{\checkmark}}& \textbf{\textcolor{teal}{\checkmark}}& 45.47 \tiny\textcolor{red}{-11.59} & 5.88 \tiny\textcolor{red}{+0.27} & 75.30 \tiny\textcolor{red}{+2.82} \\
\hline\hline
\end{tabular}
\label{ablation}
\vspace{-1em}
\end{table}

\noindent \textbf{Quantitative results.} We quantitatively compare ChangeBridge as shown in Table~\ref{tab_1}. Following standard image-quality evaluation protocols, the performance is assessed using FID~\cite{heusel2017gans} and IS~\cite{salimans2016improved} across four datasets. For semantic consistency evaluation, we use SegFormer-based~\cite{segformer} mIoU/IoU for semantic-mask and instance-layout conditions, and CLIP-based cosine similarity (CosSim) for coordinate texts. 
The first row in each table reports the results of real data, as an upper bound. Note that FID scores in remote sensing are generally higher than those in natural-image domains due to the feature distribution gap~\cite{changen,changen2}.

ChangeBridge demonstrates outstanding performance with its  variants across multiple conditional settings.  
\textit{1) Coordinate Text:} It significantly outperforms Instruct-Imagen, achieving 31.45 (-16.72) FID, 5.14 (+1.44) IS, and 0.85 (+0.04) CosSim. These results highlight its superior cross-temporal generation quality and precise adherence to the provided coordinate texts.  
\textit{2) Instance Layout:} When Evaluated on both WHU-CD and S2Looking datasets, ChangeBridge shows clear improvements. On S2Looking, it surpasses ControlNet+IPA with a 56.42 (-41.25) FID. Consistent with the visual results, ChangeBridge produces more realistic structures from pre-event inputs and matches the given spatial layouts more accurately.  
\textit{3) Semantic Mask:} ChangeBridge outperforms the state-of-the-art Changen2, achieving 59.33 (-10.10) FID and 74.26\% (+1.06\%) mIoU, showing stronger coherence and class-consistency in evolution.

\begin{figure*}
	\centering
	\includegraphics[width=\linewidth]{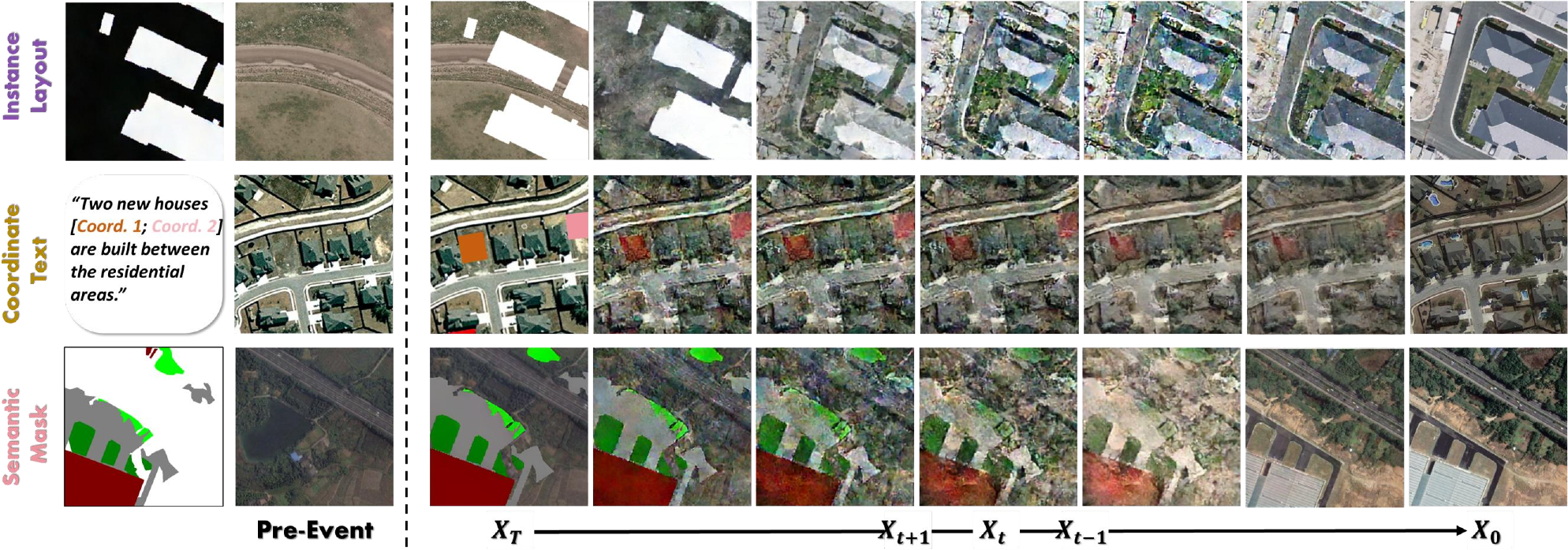}
      \vspace{-2em}
	\caption{\textbf{Intermediate sampling process} of ChangeBridge. Our diffusion bridge starts from the composed pre-event state, and ends with the post-event state. Zooming in provides better visualization.}
	\label{experiment3}
      \vspace{-0.5em}
\end{figure*}

\subsection{Ablation Study and Visualization Analysis}
\noindent \textbf{Effectiveness of components.} We conduct ablation studies to validate the effectiveness of the main components of our model. As shown in Table~\ref{ablation}, the introduction of \textit{Composed Bridge Initialization} (CompBridge) significantly improves the SD1.5 baseline, reducing the FID by 20.57. AsyDrift continues to contribute to improvements in consistency. Additionally, it is notable that before the introduction of \textit{Drift-Aware Denoising} (DriftDenoise), the FID slightly increased by 0.82, although it enhanced IS (+0.18) and IoU (+0.61\%). This demonstrates the importance of embedding drift information in the reverse denoising process.

Visualizations also show significant improvements. As shown in Figure~\ref{experiment3-1}, the baseline method, SD1.5, shows limitations in both spatial control and spatiotemporal evolution (e.g., lighting transitions in the background). In contrast, our approach shows key advancements: 1) CompBridge aligns with the instance layout, generating a building with high accuracy. 2) AsyDrift mitigates cross-temporal inconsistencies, such as the disappearance of trees (highlighted in the yellow box). 3) Finally, DriftDenoise produces high-fidelity images, significantly aiding the reverse process and enhancing overall visual quality.

\noindent \textbf{Visualization of inference evolution.} We visualize the intermediate inference process of the drift-asynchronous diffusion bridge. This process shows how the model moves from current observations and multimodal controls to the synthesized post-event images. As shown in Figure~\ref{experiment3}, the model preserves spatial and semantic consistency while evolving spatiotemporal dynamics. The foreground, like new buildings, evolves quickly and aligns with the layout, while the background changes more slowly, ensuring consistency with the pre-event image. This demonstrates that our asynchronous drift mechanism allows the foreground and background to evolve at different rates.

\begin{table}
\label{ratio}
\setlength{\tabcolsep}{3pt}
\renewcommand{\arraystretch}{1.2}
\centering
\caption{\textbf{Performance on downstream change detection tasks}. BCD denotes binary change detection, SCD refers to semantic change detection, and CC is change captioning. IoU (\%), mIoU (\%), and CIDEr-D are reported.}
\begin{tabular}{cccccc} 
\hline\hline
\multirow{2}{*}{\textbf{Data}} & \multicolumn{2}{c}{\textcolor{ForestGreen}{\textbf{BCD task}}} & \textcolor{Blue}{\textbf{SCD task}} & \textcolor{Orange}{\textbf{CC task}} \\ 
 & WHU-CD & S2Looking & SECOND & LEVIR-CC \\ 
\hline
$\mathcal{B}$ & 72.39 & 46.94 & 73.33 & 134.12 \\
$\mathcal{B} \cup \mathcal{B}'$ & 73.21\tiny\textcolor{red}{+0.82} & \textbf{48.11}\tiny\textcolor{red}{+1.17} & \textbf{74.02}\tiny\textcolor{red}{+0.69} & 138.50\tiny\textcolor{red}{+4.38} \\
$\mathcal{B} \cup 2\mathcal{B}'$ & \textbf{74.65}\tiny\textcolor{red}{+2.26} & 48.03\tiny\textcolor{red}{+1.09} & 73.57\tiny\textcolor{red}{+0.24} & \textbf{145.09}\tiny\textcolor{red}{+10.97} \\
$\mathcal{B} \cup 3\mathcal{B}'$ & 73.40\tiny\textcolor{red}{+1.01} & 47.45\tiny\textcolor{red}{+0.51} & 73.66\tiny\textcolor{red}{+0.33} & 142.31\tiny\textcolor{red}{+8.19} \\ 
\hline\hline
\end{tabular}
\vspace{-1.5em}
\end{table}

\subsection{Evaluation on Change Detection Tasks}
In this section, we evaluate the effectiveness of ChangeBridge as a data engine for three change detection tasks, using image augmentation during training. We apply different augmentation scales by varying the ratio between the original dataset \( \mathcal{B} \) and the synthesized dataset \( \mathcal{B}' \). For each task, we use the following synthesized data: instance-layout controls for binary change detection (BCD), semantic-mask samples for semantic change detection (SCD), and coordinate-text samples for change captioning (CC).

\noindent \textbf{Binary change detection.} Using BiT~\cite{Chen2021bit} on the WHU and S2Looking datasets, we evaluate with IoU. On WHU-CD, adding synthetic data improves IoU from 72.39\% to 74.65\% with a 2:1 synthetic-to-original ratio (\( \mathcal{B} \cup 2\mathcal{B}' \)), showing the benefits of synthetic data. On S2Looking, the optimal IoU of 48.11\% is achieved with a 1:1 ratio (\( \mathcal{B} \cup \mathcal{B}' \)), resulting in a 1.17\% gain over the baseline.

\noindent \textbf{Semantic change detection.} For this task, we use MambaSCD~\cite{10565926} on the SECOND dataset, evaluating with mIoU. Adding an equal amount of synthetic data (\( \mathcal{B} \cup \mathcal{B}' \)) improves mIoU slightly, from 73.33\% to 74.02\%, indicating improved semantic discrimination. However, further increases in the synthetic data ratio (beyond 1:1) lead to diminishing returns, with mIoU reaching 73.57\% at the 2:1 ratio, suggesting potential overfitting.

\noindent \textbf{Change captioning.} Using RSICCformer~\cite{9934924} on the LEVIR-CC dataset and evaluating with CIDEr-D~\cite{VedantamZP15}, the results show consistent improvement in CIDEr-D as synthetic data increases. The highest CIDEr-D score of 145.09 is achieved with a 2:1 ratio (\( \mathcal{B} \cup 2\mathcal{B}' \)), demonstrating that synthetic data enhances text-image alignment and improves caption generation. Overall, moderate augmentation scales (1:1 or 2:1 ratios) yield the best results across tasks, showing that synthetic data from ChangeBridge improves generalization in change detection.

\section{Conclusion}
We propose ChangeBridge, a conditional spatiotemporal generative model that generates realistic post-event scenarios from pre-event observations and multimodal controls. The model uses a drift-asynchronous spatiotemporal diffusion bridge. Experiments on four datasets and six baselines show that ChangeBridge achieves high-fidelity event synthesis. As a data engine, it also improves downstream change detection performance. This framework holds promising potential for applications in land-use planning and data-driven change analysis. In future work, we aim to extend ChangeBridge with a flow-matching formulation for more efficient and stable spatiotemporal generation.


\section*{Acknowledgment}
This work is supported by the National Natural Science Foundation of China (T2122014, 62272375, and 624B2109), the National Key Research and Development Program of China (2022YFB3903300), and the Key Technology Research Project of China National Petroleum Corporation (2025ZG82). The numerical calculations in this paper have been done on the supercomputing system in the Supercomputing Center of Wuhan University.




{
    \bibliographystyle{ieeenat_fullname}
    \bibliography{reference}
}




\end{document}